\documentclass[lettersize,journal]{IEEEtran}
\usepackage{graphicx}
\usepackage{cite}
\usepackage{picinpar}
\usepackage{amsmath}
\usepackage{url}
\usepackage{flushend}
\usepackage[latin1]{inputenc}
\usepackage{colortbl}
\usepackage{soul}
\usepackage{multirow}
\usepackage{pifont}
\usepackage{color}
\usepackage{alltt}
\usepackage[hidelinks]{hyperref}
\usepackage{enumerate}
\usepackage{breakurl}
\usepackage{epstopdf}
\usepackage{pbox}
\usepackage{amsmath,amssymb,amsfonts}
\usepackage{algorithmic}
\usepackage{algorithm}
\usepackage{textcomp}
\usepackage{xcolor}
\usepackage{colortbl}
\usepackage{colortbl} 
\usepackage[normalem]{ulem}
\usepackage{booktabs}
\usepackage{threeparttable}
\usepackage{changepage}
\usepackage{makecell}
\usepackage{tabularray}
\usepackage{amsmath}
\usepackage{multicol}
\usepackage{cuted}
\usepackage[normalem]{ulem}
\usepackage{caption}
\usepackage[caption=false,font=normalsize,labelfont=sf,textfont=sf]{subfig}
\begin{document}

\title{Dynamic Initialization for LiDAR-inertial SLAM}

\author{\vskip 1em
	Jie Xu$^{*}$, Yongxin Ma$^{*}$, Yixuan Li, Xuanxuan Zhang, Jun Zhou$^{\dag}$, Shenghai Yuan$^{\dag}$, and Lihua Xie, \emph{Fellow, IEEE} 
	\thanks{$^{*}$ denotes equal contribution, $^{\dag}$ denotes corresponding author.}

	\thanks{Jie Xu, Yongxin Ma and Jun Zhou are with school of Mechanical Engineering, Shandong University, Jinan 250061, China and Key Laboratory of High Efficiency and Clean Mechanical Manufacture, Ministry of Education, Jinan 250061, China (e-mail: jeff\_xu\_0503@foxmail.com; yxma@mail.sdu.edu.cn; zhoujun@sdu.edu.cn).}

	\thanks{Jie Xu, Shenghai Yuan, and Lihua Xie are with the School of Electrical and Electronic Engineering, Nanyang Technological University, 639798, Singapore (e-mail: jeff\_xu\_0503@foxmail.com; shyuan@ntu.edu.sg; elhxie@ntu.edu.sg).}

    \thanks{Yixuan Li is with the Institute of Artificial Intelligence and Robotics, Xi'an Jiaotong University, 710049, China (e-mail: 2191312118@stu.xjtu.edu.cn).}

	\thanks{Xuanxuan Zhang is with the State Key Laboratory of Information Engineering in Surveying, Mapping and Remote Sensing, Wuhan University, 430072, China (email: xuanxuanzhang@whu.edu.cn).}
 
	\thanks{\textsuperscript{1} https://github.com/lian-yue0515/D-LI-Init}
}

% The paper headers
%\markboth{Journal of \LaTeX\ Class Files,~Vol.~14, No.~8, August~2021}%
%{Shell \MakeLowercase{\textit{et al.}}: A Sample Article Using IEEEtran.cls for IEEE Journals}

%\IEEEpubid{0000--0000/00\$00.00~\copyright~2021 IEEE}
% Remember, if you use this you must call \IEEEpubidadjcol in the second
% column for its text to clear the IEEEpubid mark.

\maketitle

\begin{abstract}
The accuracy of the initial state, including initial velocity, gravity direction, and IMU biases, is critical for the initialization of LiDAR-inertial SLAM systems. Inaccurate initial values can reduce initialization speed or lead to failure. When the system faces urgent tasks, robust and fast initialization is required while the robot is moving, such as during the swift assessment of rescue environments after natural disasters, bomb disposal, and restarting LiDAR-inertial SLAM in rescue missions. However, existing initialization methods usually require the platform to remain stationary, which is ineffective when the robot is in motion. To address this issue, this paper introduces a robust and fast dynamic initialization method for LiDAR-inertial systems (D-LI-Init). This method iteratively aligns LiDAR-based odometry with IMU measurements to achieve system initialization. To enhance the reliability of the LiDAR odometry module, the LiDAR and gyroscope are tightly integrated within the ESIKF framework. The gyroscope compensates for rotational distortion in the point cloud. Translational distortion compensation occurs during the iterative update phase, resulting in the output of LiDAR-gyroscope odometry. The proposed method can initialize the system no matter the robot is moving or stationary. Experiments on public datasets and real-world environments demonstrate that the D-LI-Init algorithm can effectively serve various platforms, including vehicles, handheld devices, and UAVs. D-LI-Init completes dynamic initialization regardless of specific motion patterns. To benefit the research community, we have open-sourced our code and test datasets on GitHub\textsuperscript{1}.
\end{abstract}

\begin{IEEEkeywords}
Dynamic initialization, LiDAR-inertial, LiDAR-gyroscope odometry, simultaneous location and mapping.
\end{IEEEkeywords}

\section{Introduction}
\begin{figure}[!t]
	\centering
	\includegraphics[width=3.4in]{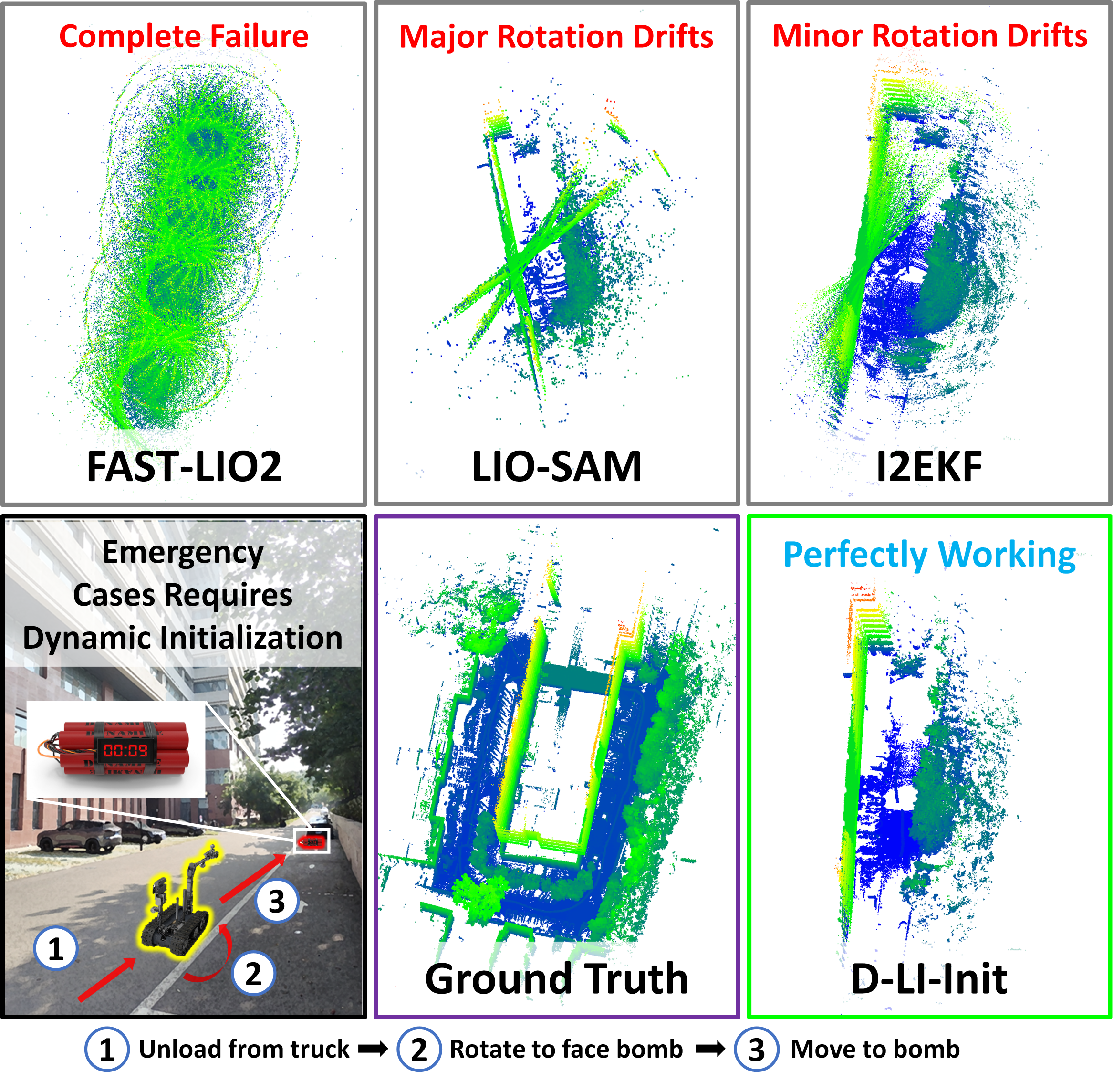}
	\caption{Demonstration of real-world scenarios where the robot requires dynamic initialization, along with point cloud maps created using SOTA algorithms and ours.}
	\label{fig_1}
	%\vspace{-12pt}
\end{figure}

LiDAR-Inertial Odometry (LIO) \cite{10517905,ji2024lio,LOG-LIO2} is a key technology in modern field robotics, but its reliance on stationary initialization poses challenges in urgent scenarios like restarting LIO in highway autonomous driving conditions \cite{MM-LINS} or bomb disposal robot rushing to area of interest. The key issue is that the IMU mitigates motion distortion in LiDAR point clouds through high-frequency motion data, improving SLAM accuracy. However, successful initialization depends on precise estimates of velocity, gravity direction, and IMU biases. Notably, the gravity direction represents a unit vector indicating the direction of gravity. Inaccurate initial estimates, especially for gravity direction and velocity, prevent the IMU from correcting motion distortion, potentially leading to faulty point cloud registration and system errors, as shown in Fig. \ref{fig_1}. \IEEEpubidadjcol

Existing research on LIO initialization \cite {FAST-LIO,Faster-LIO} is limited, primarily focusing on estimating the initial state by keeping the robot stationary for a period of time. Using those methods while the robot is in motion can result in incorrect initial estimates. This motivates the development of a robust dynamic initialization method for LIO. We hypothesize that without known initial values, the system cannot perform SLAM properly. This requires an analysis of the relationship between LiDAR odometry and IMU measurements to solve for the initial values. However, this approach poses several challenges, including (1) achieving highly accurate LiDAR odometry, (2) managing the data from both LiDAR odometry and IMU measurements to accurately solve for the initial values, and (3) ensuring that the initialization method is not constrained by specific motion patterns.
 
This paper presents D-LI-Init, a novel method that aligns LiDAR odometry with IMU pre-integration for accurate initial state estimation in both stationary and moving robots.
Enhancing the robustness and accuracy of LiDAR odometry is crucial for the effectiveness of this method. 
When the initial state, like gravity direction or velocity, is unknown, accelerometer data becomes unreliable due to interference from gravity. 
In contrast, the gyroscope is insensitive to initial state uncertainties, allowing its observational data to be correctly employed. 
As a result, LiDAR-only odometry (LO) is upgraded to LiDAR-gyroscope odometry (LGO) by tightly integrating gyroscope and LiDAR data through the error state iterated Kalman filter (ESIKF) \cite{FAST-LIO,FAST-LIO2}, improving overall accuracy and performance.
This approach produces highly accurate LiDAR odometry outputs and improves the precision of initial value estimation during the alignment process.
Subsequently, the rough initial state is integrated into the LIO system to obtain more precise odometry outputs, which are iteratively refined using IMU measurements, ultimately resulting in a high-precision initial state.
Proposed D-LI-Init has been incorporated into FAST-LIO, and experimental results demonstrate its effectiveness in accurately estimating initial values during robot motion, thereby ensuring precise subsequent state estimations (Fig. \ref{fig_1}).
The main contributions of this paper are as follows:
\begin{itemize}
	\item We propose a dynamic initialization method for LiDAR-inertial SLAM systems. This method iteratively aligns LiDAR-based odometry with IMU measurements, enabling fast and robust estimation of the system's initial state without requiring specific motion patterns.
	\item A tightly coupled integration of LiDAR and gyroscope is achieved using the ESIKF framework. The gyroscope compensates for rotational distortion in point clouds, while translational distortion is addressed during the iterative update process, resulting in improved point cloud quality and enhanced precision in state estimation. Experimental results show that this method outperforms LO in robustness and accuracy.
	\item We evaluate the dynamic initialization performance on public datasets and in real-world environments across various platforms, including vehicles, handheld devices, and UAVs. The experimental results demonstrate that the proposed method excels in dynamic initialization across various scenarios.
	\item We will open-source the code and test datasets, providing a benchmark for future research in dynamic initialization. This is the first open-source package for dynamic initialization in LiDAR-inertial SLAM systems.
\end{itemize}

\section{RELATED WORKS}
Robust initialization is a critical challenge in SLAM systems \cite{10218742,nguyen2021viral,10552105}, particularly for LiDAR-inertial SLAM, which relies on estimating initial variables through analyzing the relationship between LiDAR odometry and inertial data integration. This section explores existing work on initialization methods and LiDAR odometry.

\begin{figure*}[!t]
	\centering
	\includegraphics[width=7in]{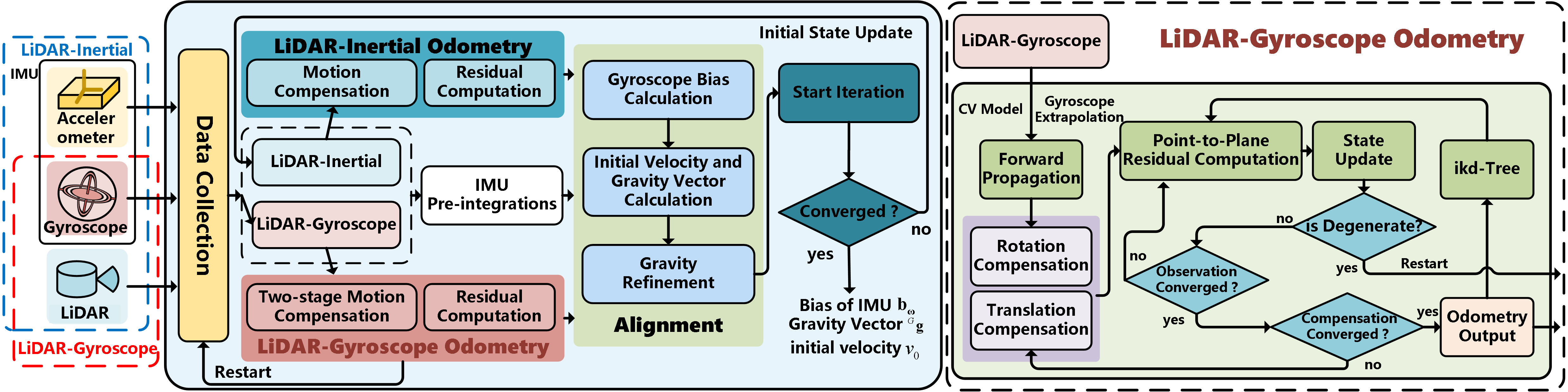}
	\caption{D-LI-Init overview.}
	\label{fig_2}
    %\vspace{-12pt}
\end{figure*}

\subsection{Initialization Methods}
In both LiDAR and visual-inertial SLAM, key initial variables such as velocity, gravity direction, and IMU biases are equally crucial for accurate system performance. \cite{VINS-initialization}. 
Existing LiDAR-inertial SLAM studies primarily rely on stationary initialization approaches \cite{LIO-SAM,iG-LIO,FAST-LIO,9197567}, where the system must remain static for a period. During this time, gyroscope biases are estimated based on the average angular velocity of the IMU, while the difference between average acceleration and local gravitational acceleration is used to determine the accelerometer bias. The gravity direction is inferred from this vector. While this stationary approach works well for vehicle-mounted systems starting from rest, it fails to accommodate systems that need to initialize while in motion, such as emergency response robots and wearable LiDAR SLAM. This limitation highlights the need for more robust dynamic initialization methods that can handle scenarios where the system is moving. 

In visual-inertial SLAM, a monocular camera lacks metric scale, requiring the robot to maintain non-zero acceleration for initialization. This led to the development of dynamic methods \cite{hhuang2024photoslam,9197334,8613749}, such as a tightly coupled closed-form solution \cite{Martinelli-591} that estimates initial state variables and feature depth using visual features and accelerometer data, though at a high computational cost.
ORB-SLAM3 \cite{ORB-SLAM3} handles IMU initialization as a maximum a posteriori estimation, using a three-step process to estimate initial parameters in dynamic situations without requiring a dedicated initialization phase. 
He et al. \cite{10205123} introduces a decoupled visual-inertial odometry initialization method, which separates the estimation of rotation and translation. This method achieves higher efficiency and robustness by solving for initial velocity and gravity direction using linear translational constraints in a globally optimal manner. 
VINS-Fusion \cite{VINS-initialization, VINS-Mono} aligns vision-only structure from motion with IMU pre-integration measurements \cite{6092505} to recover the metric scale, velocity, gravity direction, and gyroscope biases. 
While effective for visual SLAM, these VIO initialization methods are not directly applicable to LIDAR-Inerial Systems, which do not require scale recovery or motion excitation in all directions. 
However, they provide insights for developing dynamic initialization in LiDAR-inertial SLAM, where initial state estimation can be achieved by analyzing the relationship between LiDAR odometry and inertial data.

\subsection{LiDAR Odometry}
For the dynamic initialization of LiDAR-inertial SLAM systems, the accuracy of the LiDAR odometry significantly influences the precision of the initial state estimation. 
LOAM \cite{LOAM} is a classic representation of LiDAR odometry, which achieves improved registration accuracy and algorithm efficiency by extracting line and plane features for registration. However, these methods rely on a constant velocity model for motion distortion compensation, which limits their performance in real-world environments, especially under challenging motion scenarios. 
CT-ICP \cite{CT-ICP}  introduces the concept of continuous-time sensor motion, modeling it as a continuous function. It determines the vehicle's state at the beginning and end of each LiDAR frame. CT-ICP demonstrates high accuracy but comes with higher computational costs due to the increased dimensionality of the state variables and occasionally has overfitting issues. 
KISS-ICP \cite{KISS-ICP} provides a lightweight and efficient LiDAR odometry system, which estimates states with point-to-point ICP, ensuring stable operation across different environments and motion patterns. However, its effectiveness drops with sparse point clouds generated from devices like Livox Mid360. 
I2EKF-LO \cite{I2EKF-LO} introduces a dual-iterative extended Kalman filter for odometry estimation. It dynamically adjusts process noise during state estimation and establishes motion models for different sensor platforms, resulting in accurate and efficient state estimation. 
While existing systems perform reasonably well, their state estimation accuracy is slightly lower than the LiDAR-inertial systems.

For LiDAR-inertial systems \cite{FAST-LIO,FAST-LIO2}, which utilizes IMU data to assist in point cloud motion distortion compensation and tightly couples IMU and LiDAR measurements using the ESIKF, high-frequency and high-precision odometry can be output. However, for the LiDAR odometry required in dynamic initialization systems, the initial velocity and gravity direction are unknown, rendering the accelerometer in the IMU unusable. Given that the gyroscope is unaffected by initial values, using ESIKF to achieve tight coupling between the gyroscope and LiDAR could be a novel approach to improving LiDAR odometry accuracy.

\section{METHODOLOGY}
In this paper, we present a robust and fast dynamic initialization method for LiDAR-inertial SLAM, dubbed D-LI-Init. The framework, shown in Fig. \ref{fig_2}, assumes known extrinsic parameters between sensors, does not necessitate specific motion patterns from the robot and can be adapted to various platforms. To estimate the initial state, an iterative alignment between LiDAR odometry and IMU pre-integration is performed using least squares method (LSM).
\subsection{Problem Statement}
We define the global frame $^G(\cdot )$ at the initial pose of the LiDAR. The external parameter from LiDAR frame to IMU frame $^I(\cdot )$ is assumed to be known as $^I{{\bf{T}}}_L = (^I{{\bf{R}}}_L, ^I{{\bf{t}}}_L) \in SE(3)$. Let $^G{{\bf{P}}_k}{(^G}{{\bf{R}}_k}{,^G}{{\bf{t}}_k})$ represent the corresponding pose at the $k_{th}$ frame, and let the corresponding IMU pre-integration be denoted by ${\bf{\alpha }}_{k + 1}^{k},{\bf{\beta }}_{k + 1}^{k},{\bf{\gamma }}_{k + 1}^{k}$. The objective of this work is to align the LiDAR odometry poses with the IMU pre-integration by solving for the initial state $ \mathbf{x}_{\mathbf{I}}  = ({{\bf{b}}_{\boldsymbol{\omega }}}, {}^G{\bf{v}}_0, {}^G{\bf{g}})$ using LSM, where ${{\bf{b}}_{\boldsymbol{\omega }}}$, ${}^G{\bf{v}}_0$, and ${}^G{\bf{g}}$ represent the gyroscope bias, initial velocity, and gravity direction, respectively.

\subsection{LiDAR-gyroscope Odometry}
LGO module uses gyroscope data to predict rotational motion, thereby enabling compensation for rotational motion distortion during scans. A constant velocity model is employed to predict translational motion and translational motion distortion compensation is integrated into the iterative update process to enhance point cloud quality. To address the mismatch between the constant velocity model and actual motion during translational prediction, a frame segmentation of the incoming point cloud is adopted \cite{10802314, 9982225}. This approach minimizes errors caused by the constant velocity assumption, ensuring high-precision state estimation comparable to LIO. It improves the accuracy of dynamic initialization by providing better initial values for iterative alignment, reducing the required iterations, and enhancing the correctness of the D-LI-Init.

The discrete model is as follows:
\begin{equation}
	{{\mathbf{x}}_{k+1}}={{\mathbf{x}}_{k}}\boxplus \left( \Delta t\mathbf{f}\left( {{\mathbf{x}}_{k}},{{\mathbf{u}}_{k}},{{\mathbf{w}}_{k}} \right) \right),
	\label{eq_1}
\end{equation}
where $\Delta t$ is the time interval between two scans,  $\boxplus$ denotes the ``plus'' on the state manifold. The state vector $\bf{x}$, input $\bf{u}$, noise $\bf{w}$, and discrete state transfer function $\bf{f}$ are defined as:
\begin{figure*}[!t]
	\centering
	\includegraphics[width=6.1in]{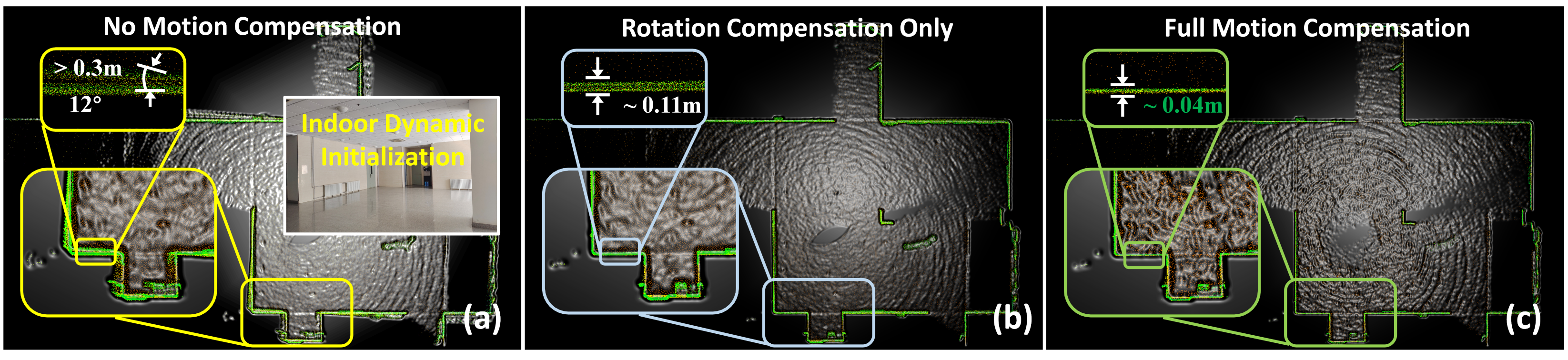}
	\caption{The effect of the motion distortion compensation method is demonstrated for a simple indoor environment. }
	\label{fig_3}
    %\vspace{-15pt}
\end{figure*}
\begin{equation}
	\begin{array}{l}
		\mathbf{x} = \left[ \begin{array}{c}
			{^G\mathbf{R}} \\
			{^G\mathbf{t}} \\
			{^G\mathbf{v}} \\
			\mathbf{b}_{\boldsymbol{\omega}}
		\end{array} \right], \quad
		\mathbf{u} = \left[ \boldsymbol{\omega}_m \right], \quad
		\mathbf{w} = \left[ \begin{array}{c}
			\mathbf{n}_{\boldsymbol{\omega}} \\
			\mathbf{n}_{\mathbf{v}} \\
			\mathbf{n}_{\mathbf{b} \boldsymbol{\omega}}
		\end{array} \right], \\
		\mathbf{f}(\mathbf{x}, \mathbf{u}, \mathbf{w}) = \left[ \begin{array}{c}
			\boldsymbol{\omega}_{m_k} - \mathbf{b}_{\boldsymbol{\omega}_k} - \mathbf{n}_{\boldsymbol{\omega}_k} \\
			{^G\mathbf{v}_k} \\
			\mathbf{n}_{\mathbf{v}_k} \\
			\mathbf{n}_{\mathbf{b} \boldsymbol{\omega}_k}
		\end{array} \right],
	\end{array}
	\label{eq_2}
\end{equation}
where $^G{\mathbf{R}} \in SO(3)$, $^G{\mathbf{t}}$ are the attitude and position in the global coordinate system (aligned with the first LiDAR frame), respectively, and $^G{\mathbf{v}}$ is the velocity in the global coordinate system, which is modeled as a random walk process driven by Gaussian noise as $\mathbf{n}_{\mathbf{v}}$. $\mathbf{b}_{\boldsymbol{\omega}}$ is the gyroscope biases, which is modeled as a random walk process driven by Gaussian noise as $\mathbf{n}_{\mathbf{b}_{\boldsymbol{\omega}}}$. $\boldsymbol{\omega}_m$ is the gyroscope measurement, and $\mathbf{n}_{\boldsymbol{\omega}}$ is the white noise measured by the IMU.

The measurement model is directly constructed by calculating point-to-plane distances, eliminating the need for feature extraction from the original point cloud, as follows:
\begin{equation}
	0 = {{\bf{h}}_j}\left( {{{\bf{x}}_k},{\bf{n}}_j^L} \right) \buildrel \Delta \over = {\bf{u}}_j^T\left( {{}^G{{\bf{R}}_k}\left( {{\bf{p}}_j^L + {\bf{n}}_j^L} \right){ + ^G}{{\bf{t}}_k} - {}^G{{\bf{q}}_j}} \right),
	\label{eq_3}
\end{equation}
where ${\bf{n}}_j^L$ is the measurement noise of LiDAR, and ${\bf{p}}_j^L$ is the coordinate of a point in the LiDAR frame. ${\bf{u}}_j^T$ is the normal vector of the plane matching the point in the mapping, and ${}^G{\bf{q}}_j$ is a point on this plane.

\subsubsection{Forward Propagation} 
The forward propagation is performed upon receiving gyroscope measurements. More specifically, by setting the process noise to zero, the state and covariance are propagated as follows:
\begin{equation}
	{\hat{\bf{x}}_{k + 1}} = {{\bf{\bar x}}_k} \boxplus \left( {\Delta t{\bf{f}}\left( {{{{\bf{\bar x}}}_k},{{\bf{u}}_k},{\bf{0}}} \right)} \right),
	\label{eq_4}
\end{equation}
\begin{equation}
	{\hat {\bf{P}}_{k + 1}} = {{\bf{F}}_{\tilde{\bf{x}}}}{\bar {\bf{P}} _k}{\bf{F}}_{{\bf{\tilde x}}}^T + {{\bf{F}}_{\bf{w}}}{\bf{QF}}_{\bf{w}}^T,
	\label{eq_5}
\end{equation}
where, $\bar{\bf{x}}$ and $\bar{\bf{P}}$ represent the posterior state and covariance matrix of the frame, respectively, while $\hat{\bf{x}}$ and $\hat{\bf{P}}$ denote the predicted state and covariance matrix, $\tilde{\bf{x}}$ represents the error state. The matrix $\bf{Q}$ is defined as the covariance matrix of $\bf{w}$. The matrices ${{\bf{F}}_{\tilde{\bf{x}}}}$ are defined as follows:
\begin{align*}
    \left[\begin{array}{cccc}
    \operatorname{Exp}\left(-\left(\boldsymbol{\omega}_{m_k}-\mathbf{b}_{\mathbf{\omega}_k}\right) \Delta t\right) & \mathbf{0}_{3 \times 3} & \mathbf{0}_{3 \times 3} & -\mathbf{I}_{3 \times 3} \Delta t \\
    \mathbf{0}_{3 \times 3} & \mathbf{I}_{3 \times 3} & \mathbf{I}_{3 \times 3} \Delta t & \mathbf{0}_{3 \times 3} \\
    \mathbf{0}_{3 \times 3} & \mathbf{0}_{3 \times 3} & \mathbf{I}_{3 \times 3} & \mathbf{0}_{3 \times 3} \\
    \mathbf{0}_{3 \times 3} & \mathbf{0}_{3 \times 3} & \mathbf{0}_{3 \times 3} & \mathbf{I}_{3 \times 3}
    \end{array}\right], 
\end{align*}
    % \mathbf{F}_{\tilde{\bf{x}}} = 
and $\mathbf{F}_{\mathbf{w}}$ is defined as:
\begin{align*}
   \left[\begin{array}{ccc}
    \mathbf{I}_{3 \times 3} \Delta t & \mathbf{0}_{3 \times 3} & \mathbf{0}_{3 \times 3} \\
    \mathbf{0}_{3 \times 3} & \mathbf{0}_{3 \times 3} & \mathbf{0}_{3 \times 3} \\
    \mathbf{0}_{3 \times 3} & \mathbf{I}_{3 \times 3} \Delta t & \mathbf{0}_{3 \times 3} \\
    \mathbf{0}_{3 \times 3} & \mathbf{0}_{3 \times 3} & \mathbf{I}_{3 \times 3} \Delta t
    \end{array}\right].
\end{align*}

\subsubsection{Backward Propagation and Rotational Distortion Compensation} 
The backward propagation process is primarily used to compute the pose of each point within the LiDAR frame. This process focuses solely on the rotational component, denoted as ${{\bf{\hat x}}_{k - 1}} = {{\bf{\hat x}}_k} \boxplus \left( { - \Delta t{\bf{f}}\left( {{{{\bf{\hat x}}}_k},{{\bf{u}}_k},{\bf{0}}} \right)} \right)$, which can be simplified accordingly for the purpose of removing rotational distortion. Specifically, this can be expressed as \eqref{eq_8}.

\begin{equation}
	\hat{\mathbf{R}}_{i-1}=^{{I_k}}{{{\bf{\hat R}}}_i} \operatorname{Exp}\left(\left(\hat{\boldsymbol{b}}_{\boldsymbol{\omega}_k}-\boldsymbol{\omega}_{m_{i-1}}\right) \Delta t\right)
	\label{eq_8}
\end{equation}

The process determines the relative pose between point $i$ ($i \in (k - 1, k]$) within the LiDAR frame and the end of the scan at time ${t_k}$, denoted as $^{{I_k}}{{\bf{\hat T}}_i} = \left( {^{{I_k}}{{{\bf{\hat R}}}_i},0} \right)$. This relative orientation allows for the projection of local measurements ${}^{{L_i}}{{\bf{p}}_i}$ onto the coordinate system $^{{L_i}}{{\bf{p}}_k}$ at the end of the scan, thus facilitating the transformation of all points within the LiDAR frame to the coordinate system at the frame's end.
\begin{equation}
	^{{L_i}}{{\bf{p}}_k} = \left( {{}^I{\bf{T}}_L^{ - 1}} \right)\left( {{}^{{I_k}}{{\hat{\bf{T}}}_i}} \right)\left( {^I{{\bf{T}}_L}} \right){}^{{L_i}}{{\bf{p}}_i}
	\label{eq_9}
\end{equation}
where ${^I{{\bf{T}}_L}}$ represents a known extrinsic parameter.

\subsubsection{Iteration}
Each iteration includes iterative motion distortion compensation and iterative observation update.

\textit{\romannumeral1) Iterative Motion Distortion Compensation: }To correct translational motion distortion in the system, consistent with the backward propagation process, the point cloud of the current frame $i$ ($i \in (k - 1, k]$) is uniformly transformed into the LiDAR coordinate system at the end time $t_k$.

For a point ${}^{{L_i}}{{\bf{p}}_i}$ in the LiDAR coordinate system generated at time $t_i$ within frame $k$, given the predicted state transformation ${{}^G{{\hat{\bf{T}}}_k}}$ for frame $k$ and the posterior state transformation ${}^G{\bar {\bf{T}} _{k - 1}}$ for frame $k-1$, the distortion-corrected point $^{{L_i}}{{\bf{p}}_k}$ is obtained by determining the transformation ${}^G{{\bf{T}}_i}$ of the LiDAR coordinate system in the global coordinate system at time $t_i$.

\begin{equation}
	^{{L_i}}{{\bf{p}}_k} = {\left( {{}^G{{\hat{\bf{T}}}_k}} \right)^{ - 1}}\left( {{}^G{{\hat{\bf{T}}}_i}} \right){}^{{L_i}}{{\bf{p}}_i},
	\label{eq_10}
\end{equation}

\begin{equation}
	{}^G{\hat{\bf{T}}_k} = [{\bf{I}}\begin{array}{*{20}{c}}
	\end{array}\mid \begin{array}{*{20}{c}}
	\end{array}{}^G{\hat {\bf{t}}_i}],
	\label{eq_11}
\end{equation}
where ${}^G{\hat {\bf{t}}_i}$ is predicted by linear interpolation:
\begin{equation}
	{}^G{\hat {\bf{t}}_i} = {}^G{\bar {\bf{t}} _{k - 1}} + {\mathop{\rm scale}\nolimits} \left( {{}^G{{\bf{t}}_k} - {}^G{{\bf{t}}_{k - 1}}} \right),
	\label{eq_12}
\end{equation}
\begin{equation}
	{\rm{scale}} = \frac{{{t_i} - {t_{k - 1}}}}{{{t_k} - {t_{k - 1}}}}.
	\label{eq_13}
\end{equation}

The effect of the motion distortion compensation method described in this paper is shown in Fig. \ref{fig_3}.

\begin{figure*}[!t]
	\centering
	\includegraphics[width=6.4in]{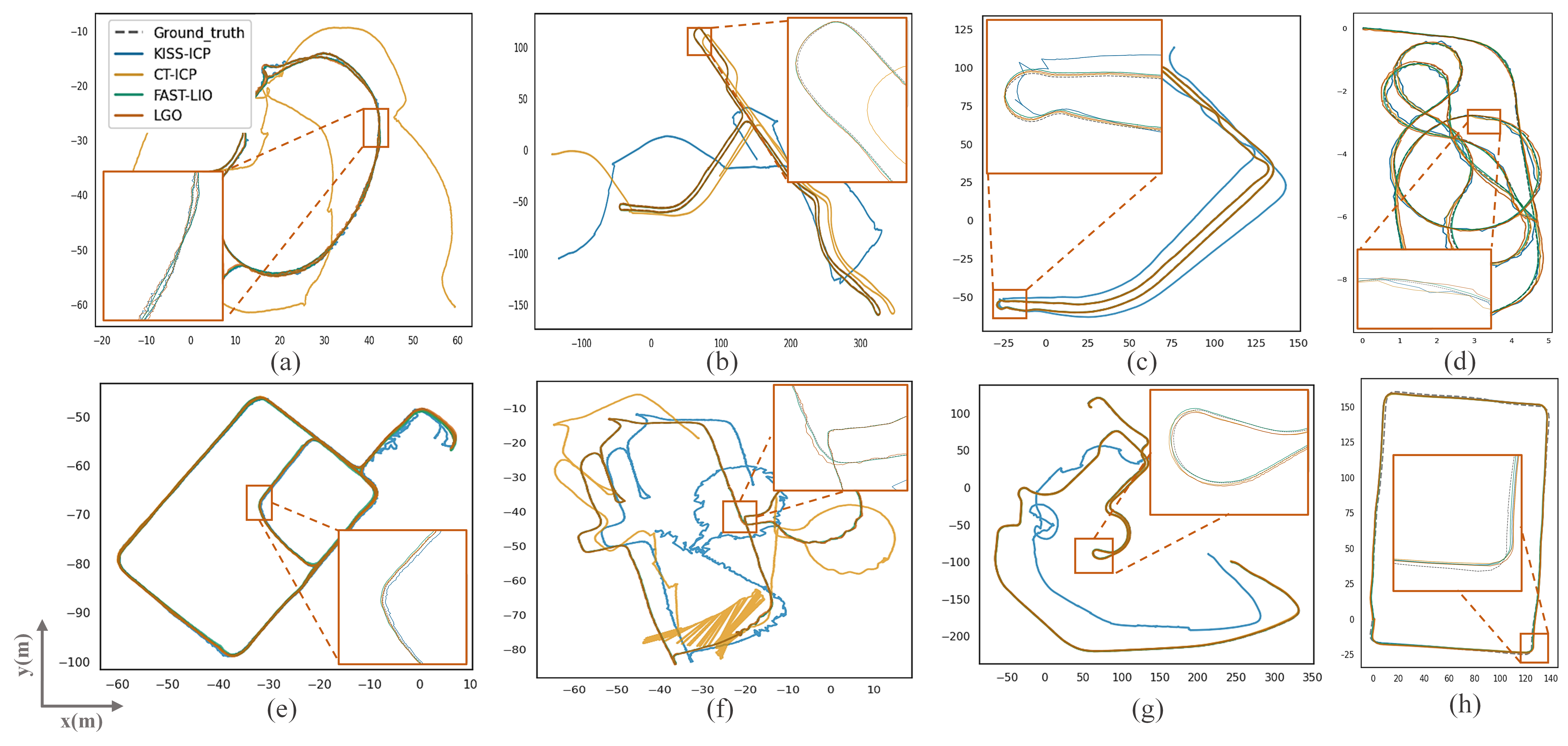}
	\caption{Trajectory generated by various algorithms for different datasets. LGO performs well, outperforming LO and approaching LIO. The subfigures depict the following: (a) - (h) trajectories of various algorithms in the NEW\_quad, MCD\_night\_04, MCD\_night\_13, Gnd\_MBLR\_1, NEW\_sloitter, 	NEW\_math, MCD\_day\_02, UbanLoco\_test, respectively.}
	\label{fig_5}
    %\vspace{-12pt}
\end{figure*}

\textit{\romannumeral2) Iterative Observation Update: }The motion-compensated scan, denoted as $\left\{ {^{{L_i}}{{\bf{p}}_k}} \right\}$, provides an implicit measurement of the predicted pose ${}^G{{\bf{T}}_{k + 1}}$, which is formulated as a point-to-plane distance residual function. Based on this, observation updates are iteratively performed within the ESIKF framework until the estimated state ${{\bf{x}}_{k - 1}}$ converges. The converged state estimate, denoted as ${\bar{\bf{x}}_{k - 1}}$, is then used to propagate subsequent IMU measurements. For more details on this iterative observation update process, refer to \cite{FAST-LIO}.

\subsubsection{Degeneracy Detection}
For the LGO module, degeneracy detection ensures that the module provides high-precision LiDAR odometry while also serving as a criterion for determining the reliability of the data. The local submatrix of the covariance matrix better accounts for the rotational and translational constraints \cite{xu2024selectivekalmanfilterfuse}. As the rotational and translational matrix blocks clearly contain information about the strength of constraints on the robot's state, this information can be utilized to establish degeneracy detection standards. The covariance matrix can thus be divided into several submatrices:
\begin{equation}
	{\bf{P}} = {\left[ {\begin{array}{*{20}{l}}
				{{{\bf{P}}_{rr}}}&{{{\bf{P}}_{rt}}}\\
				{{{\bf{P}}_{tr}}}&{{{\bf{P}}_{tt}}}
		\end{array}} \right]_{6 \times 6}} ,
	\label{eq_14}
\end{equation}
where $r$ denotes rotation, $t$ denotes translation, ${\bf{P}}_{rr}$ contains only information about the rotation variable, and ${\bf{P}}_{tt}$ contains only information about the translation variable.

The constraints can be distinctly categorized into translation and rotation. Analyzing the covariance matrix as a whole would lead to the coupling of rotation and translation. Since the scales and types of rotation and translation are different, this coupling complicates the setting of threshold parameters for degeneracy detection. Therefore, eigenvalue decomposition is performed separately for rotation and translation:
\begin{equation}
	{{\bf{P}}_{rr}} = {{\bf{V}}_r}{\Sigma _r}{\bf{V}}_r^ \top ,\quad {{\bf{P}}_{tt}} = {{\bf{V}}_t}{\Sigma _t}{\bf{V}}_t^ \top ,
	\label{eq_15}
\end{equation}
where ${\bf{V}}_r$ and ${\bf{V}}_t$ are the eigenvectors in the matrix, and ${\Sigma _r}$ and ${\Sigma _t}$ are diagonal matrices containing information about the eigenvalues of ${\bf{P}}_{rr}$ and ${\bf{P}}_{tt}$.

The eigenvalues in ${\bf{V}}_r$ and ${\bf{V}}_t$ provide a direct measure of the system's observed strength with respect to the rotation and translation constraints. By monitoring the maximum eigenvalues ${\lambda_{k}^{(\text{max})}}^r$ and ${\lambda_{k}^{(\text{max})}}^t$ corresponding to each frame's eigenvalue matrix, we can assess whether the system is experiencing degeneracy. The degeneracy thresholds $\xi ^{r}$ and ${\xi ^{t}}$ are empirically determined for this purpose. It is important to note that the relevant threshold settings are closely tied to the sensor type. Therefore, we establish the thresholds through experiments on datasets using various sensor types to ensure the robustness of the degeneracy detection mechanism.
\begin{equation}
	O_{k}= \begin{cases}1, & \text { if } {\lambda_{k}^{(\text{max})}}^r <\xi ^{r}, {\lambda_{k}^{(\text{max})}}^t<\xi ^{t} \\ 0, & \text { otherwise }\end{cases},
	\label{eq_16}
\end{equation}
where $O_{k}$ indicates whether the $k_{th}$ data system is degenerate. A value of $1$ signifies that no degeneracy has occurred, while a value of $0$ indicates system degeneracy. If degeneracy is detected, the data collected at that moment is considered unreliable, implying that the available LiDAR data do not provide accurate odometry information, which may result in initialization failure. Consequently, the previously accumulated frames are discarded, and data collection is restarted.
\begin{table*}
	\centering
	\caption{Comparison of ATE (RMSE) in Meter Across Various \textbf{Dynamic} Initialization Datasets}
        \begin{tabular}{cccccccccc} 
        \toprule
        \begin{tabular}[c]{@{}c@{}}\textbf{Sequences }\\\textbf{ ( Starting Time(s) )}\end{tabular} & \textbf{Motion Patterns} & \textbf{LiDAR Type} & \textbf{Platform}         & \textbf{FAST-LIO2}          & \textbf{LIO-SAM} & \textbf{KISS-ICP} & \textbf{CT-ICP}      & \textbf{I2EKF} & \textbf{D-LI-Init}  \\ 
        \midrule
        Gnd\_MBLR\_1 (28)                                                                           & Spinning                 & Velodyne            & \multirow{5}{*}{Vehicle}  & 2.807                       & 0.046            & 0.076             & \textbf{0.033}       & 2.864          & \uline{0.041}       \\
        Gnd\_MBLR\_3 (30)                                                                           & Spinning                 & Velodyne            &                           & 3.461                       & 1.009            & \uline{0.081}     & 0.083                & 4.252          & \textbf{0.079}      \\
        M2DGR\_street (220)                                                                         & L-R Turns                & Velodyne            &                           & 4.456                       & 0.113            & 0.230             & \uline{0.108}        & 6.875          & \textbf{0.101}      \\
        MCD\_day\_02 (112)                                                                          & Turning                  & Ouster              &                           & 3.908                       & 0.063            & 0.070             & \textbf{0.045}       & 0.114          & 0.073               \\
        MCD\_night\_04 (45)                                                                         & Turning                  & Ouster              &                           & 5.517                       & 18.753           & \uline{0.056}     & 0.115                & 0.142          & \textbf{0.011}      \\ 
        \midrule
        NTU\_spms\_01 (265)                                                                         & Spinning                 & Ouster              & \multirow{2}{*}{UAV}      & 8.665                       & 216.439          & 2.496             & -\textsuperscript{b} & \textbf{0.073} & \uline{0.317}       \\
        NTU\_rtp\_03 (235)                                                                          & Gliding                  & Ouster              &                           & $\times$\textsuperscript{a} & 230.630          & \uline{0.974}     & -                    & 2.154          & \textbf{0.092}      \\ 
        \midrule
        NEW\_math\_0 (111)                                                                          & S-S Swinging             & Ouster              & \multirow{4}{*}{Handheld} & $\times$                    & 0.038            & 0.129             & \uline{0.095}        & 0.104          & \textbf{0.036}      \\
        NEW\_math\_1 (20)                                                                           & S-S Swinging             & Ouster              &                           & $\times$                    & 0.130            & 0.133             & \uline{0.061}        & 0.363          & \textbf{0.027}      \\
        NEW\_park (112)                                                                             & Turning                  & Ouster              &                           & 1.754                       & 0.065            & 0.141             & \uline{0.084}        & 0.074          & \textbf{0.064}      \\
        R3LIVE\_seq\_1 (5)                                                                          & F-B Swinging             & AVIA                &                           & $\times$                    & 0.023            & -                 & -                    & \uline{0.030}  & \textbf{0.016}      \\
        \bottomrule
        \end{tabular}
	\vspace{1pt}
	\begin{adjustwidth}{0cm}{0cm}
		\footnotesize{\textsuperscript{a} ``$\times$'' denotes the result drifted. \textsuperscript{b}  ``--'' denotes numerical instability leading to crashing either at Start-up or at degeneracy encounter. Best results are \textbf{boldened}, and second-best results are \underline{underlined}. ``L-R Turns'' denotes Left and Right Turns, ``S-S Swinging'' denotes Side-to-Side Swinging, ``F-B Swinging'' denotes Forward and Backward Swinging, ``D-LI-Init'' denotes FAST-LIO enhanced by D-LI-Init. The table below follows the same rules.}
	\end{adjustwidth}
	\label{tab1}
    %\vspace{-12pt}
\end{table*}

\subsection{Dynamic LiDAR-inertial Initialization}
The primary objective of the D-LI-Init method is to accurately estimate the initial values, including the initial velocity, gravity direction, and IMU biases. In the dynamic initialization process, the accurate estimation of accelerometer biases requires the platform to undergo at least a 30-degree rotation to decouple them from the gravitational acceleration \cite {VINS-initialization}, which limits the general applicability of the method. Therefore, the estimation of accelerometer biases is deliberately omitted, with their determination deferred to subsequent estimation systems. The LGO module tightly couples LiDAR observations with gyroscope data to output high-precision odometry. IMU data is provided, and pre-integration is used to obtain corresponding constraint information. Additionally, the LGO module includes a degeneracy detection mechanism to determine whether the LiDAR point cloud observations have degenerated. This ensures a reliable evaluation of whether the estimated trajectory is suitable for the dynamic initialization process. Once a sufficient number of trajectory poses are available, the dynamic initialization module is initiated. Initially, a linear system is constructed using LGO data and IMU pre-integration data to solve for the initial parameters, producing a coarse set of initial values. These initial values are then fed back into the LIO system to obtain more precise odometry information, which is subsequently aligned with IMU pre-integration data. Iterative updates are performed, ultimately yielding the initial velocity, gravity direction, and gyroscope biases.

\subsubsection{Data Collection}
A sufficient amount of data is initially gathered based on the degeneracy detection criterion. In this study, 20 frames of data are selected to ensure the accuracy of the initial value estimation (This value is determined empirically and can be applied in general scenarios). The corresponding poses are obtained through the LGO module (i.e., $^G{{\bf{P}}_k}{(^G}{{\bf{R}}_k}{,^G}{{\bf{t}}_k})$), and the IMU pre-integration corresponding to the same time intervals (i.e., ${\bf{\alpha }}_{k + 1}^{k},{\bf{\beta }}_{k + 1}^{k},{\bf{\gamma }}_{k + 1}^{k}$) are simultaneously collected for subsequent calculations.

\subsubsection{Gyroscope Biases Initialization}

To solve for the gyroscope biases, we align the rotation constraints between two consecutive frames. Given the rotations $^G{{\bf{R}}_{k + 1}}$ and $^G{{\bf{R}}_{k}}$ for these frames, and the relative rotation constraints from the IMU pre-integration ${\bf{\gamma }}_{{I_{k + 1}}}^{{I_k}}$, we can construct the following linear equation to find the solution:
\begin{equation}
	{\min _{\delta {{\bf{b}}_{\boldsymbol{\omega }}}}}\sum\limits_{k = 0}^{20} {{{\left\| {^G{{\bf{R}}_{k + 1}}}^{ - 1}{ \otimes ^G}{{\bf{R}}_k} \otimes {\bf{\gamma }}_{k + 1}^{k} \right\|}^2}} ,
	\label{eq_17}
\end{equation}
\begin{equation}
	{\bf{\gamma }}_{k + 1}^{k} \approx {\bf{\hat \gamma }}_{k + 1}^{k} \otimes \left[ {\begin{array}{*{20}{c}}
			1\\
			{\frac{1}{2}{\bf{J}}_{{{\bf{b}}_{\boldsymbol{\omega }}}}^\gamma \delta {{\bf{b}}_{\boldsymbol{\omega }}}}
	\end{array}} \right],
	\label{eq_18}
\end{equation}
where ${\bf{J}}_{{{\bf{b}}_{\boldsymbol{\omega }}}}^\gamma $ denotes the Jacobi matrix of the relative change in rotation with respect to $\delta {{\bf{b}}_{\boldsymbol{\omega }}}$. The $\delta {{\bf{b}}_{\boldsymbol{\omega }}}$ solved by the above system is equal to the initial ${{\bf{b}}_{\boldsymbol{\omega }}}$. Subsequently, the initial ${{\bf{b}}_{\boldsymbol{\omega }}}$ is utilized to re-propagate the IMU pre-integration for the solution of the subsequent variables.

\subsubsection{Initial Velocity and Gravity Direction Initialization}
The initial velocity and gravity direction are estimated by aligning the velocity and translation constraints derived from the odometry data and IMU pre-integration. The variables that need to be solved in this process are as follows:
\begin{equation}
	\mathbf{x}_{\mathbb{I}} = {\left[ {{}^G{{\bf{v}}_0},{}^G{{\bf{v}}_1}, \cdots {}^G{{\bf{v}}_k} \cdots ,{}^G{{\bf{v}}_{20}},{}^G{\bf{g}}} \right]^T},
	\label{eq_19}
\end{equation}
where ${}^G{{\bf{v}}_k}$ is the velocity in LiDAR coordinates at frame $k$. ${}^G{\bf{g}}$ is the gravitational acceleration at the completion of the first LiDAR frame. $(\cdot )^T$ denotes the transpose of a matrix.
\begin{figure*}[!t]
	\centering
	\includegraphics[width=6.2in]{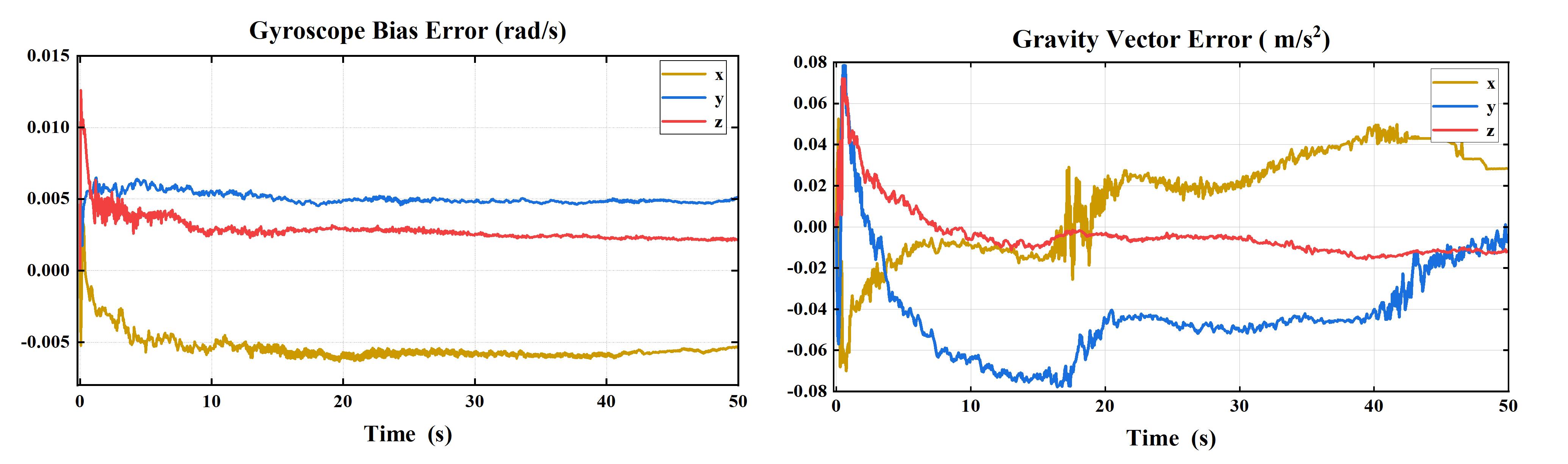}
	\caption{The error for the D-LI-Init method in estimating the gravity vector and initial gyroscope bias, where the error denotes the difference between the further refined gyroscope bias and gravity vector obtained by the LiDAR-inertial system (initialized using the D-LI-Init method) after completing dynamic initialization and the initial values estimated by the D-LI-Init method.}
	\label{fig_4}
    %\vspace{-12pt}
\end{figure*}

By constructing the state increments in the discrete form of velocity and pose between two frames, the equations for IMU pre-integration \cite{VINS-Mono} are derived as follows:
\begin{equation}
	\begin{array}{l}
		{\boldsymbol{\alpha}}_{k+1}^{k} = {^G \mathbf{R}_k}^T \left( {^I \mathbf{t}_{k+1} - ^I \mathbf{t}_k + \frac{1}{2} {^G \mathbf{g}} \Delta t^2 - {^G \mathbf{v}_k} \Delta t} \right), \\
		{\boldsymbol{\beta}}_{k+1}^{k} = {^G \mathbf{R}_k}^T \left( {^G \mathbf{R}_{k+1} {^G \mathbf{v}_{k+1}} + {^G \mathbf{g}} \Delta t - {^G \mathbf{R}_k} {^G \mathbf{v}_k}} \right)
	\end{array}
	\label{eq_20}
\end{equation}
here,  $\Delta t$ is the time interval between two frames. By combining the LiDAR data with the external parameters of the IMU (i.e., $^I{{\bf{t}}_k} = {}^G{{\bf{t}}_k}{ - ^G}{{\bf{R}}_k}^I{{\bf{t}}_L}$, where $^I{{\bf{t}}_L}$ denotes the translation external parameter), the following linear model is obtained.
\begin{equation}
	\begin{array}{l}
		\mathbf{\hat{z}}_{k+1}^{k} = \left[ \begin{array}{c}
			{\boldsymbol{\hat{\alpha}}_{k+1}^{k} - {^G \mathbf{R}_k}^T \left( {^G \Delta\mathbf{t}_{k+1}^k - {^G \mathbf{R}_{k+1}} {^I \mathbf{t}_L}} \right) - {^I \mathbf{t}_L}} \\
			{\hat{\boldsymbol{\beta}}_{k+1}^{k}}
		\end{array} \right] \\
		\qquad \; \, = \mathbf{H}_{k+1}^{k} \mathbf{x}_{\mathbb{I}_k} + \mathbf{n}_{k+1}^{k},
	\end{array}
	\label{eq_21}
\end{equation}
here, $^G \Delta\mathbf{t}_{k+1}^k  = ^G \mathbf{t}_{k+1} - ^G \mathbf{t}_k$, $	{{\mathbf{x}_{\mathbb{I}_k} }} = {\left[ {{}^G{{\bf{v}}_k},{}^G{{\bf{v}}_{k + 1}},{}^G{\bf{g}}} \right]^T}$, 
\begin{align*}
	\mathbf{H}_{k+1}^{k} = \left[ \begin{array}{ccc}
	-\mathbf{I} \Delta t & \mathbf{0} & \frac{1}{2} ({^G \mathbf{R}_k}^T) \Delta t^2 \\
	-\mathbf{I} & ({^G \mathbf{R}_k}^T) {^G \mathbf{R}_{k+1}} & {^G \mathbf{R}_k}^T \Delta t
	\end{array} \right],
\end{align*}
solve the linear least squares problem to obtain $\mathbf{x}_{\mathbb{I}}$:
\begin{equation}
	{\min _{{{\mathbf{x}_{\mathbb{I}_k} }}}}\sum\limits_{k = 0}^{20} {{{\left\| {{\bf{\hat z}}_{k + 1}^{k} - {\bf{H}}_{k + 1}^{{k}}{{\mathbf{x}_{\mathbb{I}_k} }}} \right\|}^2}} .
	\label{eq_24}
\end{equation}

The aforementioned process does not account for the fixed magnitude of gravity, as incorporating this constraint would render the system nonlinear, making it difficult to solve \cite{VINS-initialization}. To address this, the gravity norm is constrained by optimizing the two-dimensional error state in the tangent space of the gravity vector, thereby refining the gravitational acceleration. Finally, all LiDAR frames $^G(\cdot )$ are rotated into the IMU frame $^I(\cdot )$, enabling seamless integration into the LIO system for subsequent estimation.

\begin{table}
	\centering
	\caption{LGO Performance is Demonstrated through ATE (RMSE) in Meter Comparison.}
    \begin{tabular}{ccccc} 
    \toprule
    \textbf{Sequences} & \textbf{KISS-ICP} & \textbf{CT-ICP} & \textbf{FAST-LIO} & \textbf{LGO}    \\ 
    \midrule
    Gnd\_MBLR\_1       & 0.103             & 0.062           & \textbf{0.033}    & \uline{0.061}   \\
    UbanLoco\_test     & 2.183             & 1.297           & \uline{1.158}     & \textbf{1.127}  \\
    M2DGR\_gate        & 7.063             & 0.328           & \textbf{0.321}    & \uline{0.334}   \\
    NEW\_sloitter      & 1.591             & 0.150           & \textbf{0.119}    & \uline{0.144}   \\
    NEW\_math          & 9.885             & 9.811           & \textbf{0.098}    & \uline{0.211}   \\
    NEW\_quad          & 0.161             & 20.052          & \textbf{0.066}    & \uline{0.104}   \\
    MCD\_day\_02       & 11.422            & 0.476           & \uline{0.466}     & \textbf{0.396}  \\
    MCD\_night\_04     & 88.092            & 1.665           & \uline{0.862}     & \textbf{0.835}  \\
    MCD\_night\_13     & 89.269            & 55.162          & \textbf{0.873}     & \uline{0.878}  \\
%    \midrule
%    {Average}   & 23.308            & 9.889           & \textbf{0.444}             & \uline{0.454}           \\
    \bottomrule
    \end{tabular}
	\label{tab2}
\end{table}
\begin{table}
	\centering
	\caption{Error in Solving for Initial Velocity ($m/s^2$)}
	\begin{tabular}{cccc} 
		\toprule
		\textbf{Sequences}      & \textbf{Ground Truth} & \textbf{Solution Results} & \textbf{Error}  \\ 
		\midrule
		Gnd\_MBLR\_1   & 0.778        & 0.948            & 0.170  \\
		Gnd\_MBLR\_3   & 0.043        & 0.398            & 0.355  \\
		M2DGR\_street  & 0.772        & 1.740            & 0.969  \\
		MCD\_day\_02   & 1.664        & 2.663            & 0.998  \\
		MCD\_night\_04 & 18.330       & 19.422           & 1.092  \\
		NEW\_math\_0   & 2.035        & 2.029            & 0.006  \\
		NEW\_park      & 1.364        & 1.666            & 0.302  \\
		\bottomrule
	\end{tabular}
	\label{tab3}
    %\vspace{-12pt}
\end{table}

\subsubsection{Iterative Update}
To ensure the accuracy of initial value estimation, the primary objective is to achieve higher precision in LiDAR odometry output. In iterative process, the precision of the initial value estimations directly impacts the quality of the point clouds generated by the LiDAR odometry, resulting in a `chicken-egg' problem. Consequently, the coarse initial values obtained through the alignment of LGO and IMU pre-integration are utilized in the LIO system to enhance the precision of the LiDAR odometry. Repeatedly using the above method for the initial values, iterative updates are performed until convergence. For the LIO system, the accuracy of the initial velocity is critical for the quality of pose prediction and point cloud motion distortion compensation based on IMU. Therefore, the convergence of the iteration process is determined by the accuracy of the initial velocity:
\begin{equation}
	\left| {{{\left\| {{}^G{\bf{v}}_0^{(i)}} \right\|}^2} - {{\left\| {{}^G{\bf{v}}_0^{(i - 1)}} \right\|}^2}} \right| < {\Gamma _v},
	\label{eq_25}
\end{equation}
where, ${}^G{\bf{v}}_0^{(i)}$ represents the result of solving the initial velocity for the $i_{th}$ iteration, and $\Gamma _v$ denotes the threshold for the difference in the norms between the two initial velocity solutions. This threshold reflects the magnitude of change between the results of successive solutions.

\begin{table*}
	\centering
	\caption{ATE (RMSE) in Meter Comparison of Real-World Proprietary Datasets (Indoor)}
        \begin{tabular}{cccccccccc}
        \toprule
        \textbf{Sequences} & \textbf{Motion Patterns} & \textbf{Platform}                   & \textbf{Initial Velocity} & \textbf{Average Depth} & \textbf{FAST-LIO}         & \textbf{LIO-SAM} & \textbf{KISS-ICP} & \textbf{I2EKF} & \textbf{D-LI-Init}  \\ 
        \midrule
        Indoor\_1          & L-R Turns                & \multirow{14}{*}{Vehicle}  & 0.525                                     & 4.256             & $\times$                  & 0.069            & 0.058             & \uline{0.031}  & \textbf{0.018}      \\
        Indoor\_2          & Spinning                 &                            & 0.693                                     & 7.118             & $\times$                  & 11.065           & \uline{0.091}     & 0.273          & \textbf{0.016}      \\
        Indoor\_3          & L-R Turns                &                            & 1.292                                     & 7.052             & 0.087                     & \uline{0.022}    & 0.063             & 0.026          & \textbf{0.022}      \\
        Indoor\_4          & L-R Turns                &                            & 1.985                                     & 6.916             & $\times$                  & 0.046            & 0.064             & \textbf{0.032} & \uline{0.033}       \\
        Indoor\_5          & Spinning                 &                            & 0.797                                     & 5.686             & 5.530                     & \uline{0.025}    & 0.097             & 0.242          & \textbf{0.019}      \\
        Indoor\_6          & Spinning                 &                            & 0.452                                     & 7.269             & 2.548                     & \uline{0.037}    & 0.090             & 0.080          & \textbf{0.019}      \\
        Indoor\_7          & L-R Turns                &                            & 0.812                                     & 4.811             & 2.679                     & \uline{0.024}    & 0.068             & 0.120          & \textbf{0.018}      \\
        Indoor\_8          & L-R Turns                &                            & 0.953                                     & 6.192             & $\times$                  & 0.223            & 0.109             & \uline{0.080}  & \textbf{0.016}      \\
        Indoor\_9          & L-R Turns                &                            & 0.549                                     & 5.595             & 2.597                     & 0.053            & 0.142             & \uline{0.049}  & \textbf{0.011}      \\
        Indoor\_10         & L-R Turns                &                            & 1.141                                     & 4.196             & $\times$                  & \uline{0.127}    & 0.279             & 0.289          & \textbf{0.027}      \\
        Indoor\_11         & Spinning                 &                            & 0.663                                     & 6.472             & $\times$                  & \uline{0.040}    & 0.063             & $\times$       & \textbf{0.018}      \\
        Indoor\_12         & L-R Turns                &                            & 1.208                                     & 5.526             & $\times$                  & 1.209            & \uline{0.471}     & 1.200          & \textbf{0.023}      \\
        Indoor\_13         & Spinning                 &                            & 0.40                                     & 5.68             & $\times$                  & \uline{0.034}    & 0.162             & 0.093          & \textbf{0.019}      \\
        Indoor\_14         & L-R Turns                &                            & 0.609                                     & 7.266             & 4.981                     & 0.041            & 0.058             & \uline{0.027}  & \textbf{0.023}      \\ 
        \midrule
        Indoor\_15         & F-B Swinging             & \multirow{14}{*}{Handheld} & 2.135                                     & 4.311             & $\times$                  & 0.517            & \uline{0.209}     & 0.235          & \textbf{0.019}      \\
        Indoor\_16         & S-S Swinging             &                            & 1.592                                     & 4.156             & 0.151                     & 0.237            & 0.191             & \uline{0.027}  & \textbf{0.022}      \\
        Indoor\_17         & F-B Swinging             &                            & 1.887                                     & 4.363             & $\times$ & 0.126            & 0.163             & \uline{0.038}  & \textbf{0.028}      \\
        Indoor\_18         & S-S Swinging             &                            & 1.034                                     & 5.328             & $\times$                  & \uline{0.055}    & 0.155             & 0.210          & \textbf{0.024}      \\
        Indoor\_19         & S-S Swinging             &                            & 1.896                                     & 3.367             & $\times$                  & \uline{0.153}    & 0.394             & 1.379          & \textbf{0.037}      \\
        Indoor\_20         & F-B Swinging             &                            & 1.152                                     & 5.273             & $\times$                  & \uline{0.032}    & 0.113             & 0.043          & \textbf{0.017}      \\
        Indoor\_21         & S-S Swinging             &                            & 1.657                                     & 4.890             & $\times$                  & \uline{0.054}    & 0.181             & 1.147          & \textbf{0.027}      \\
        Indoor\_22         & Turning                  &                            & 1.966                                     & 3.928             & $\times$                  & 0.192            & 0.134             & \textbf{0.026} & \uline{0.028}       \\
        Indoor\_23         & Turning                  &                            & 0.802                                     & 5.266             & $\times$                  & 0.045            & 0.058             & \uline{0.029}  & \textbf{0.020}      \\
        Indoor\_24         & S-S Swinging             &                            & 0.844                                     & 4.629             & 2.322                     & \uline{0.100}    & 0.252             & 0.590          & \textbf{0.027}      \\
        Indoor\_25         & S-S Swinging             &                            & 0.676                                     & 4.083             & $\times$                  & 0.055            & 0.113             & \uline{0.035}  & \textbf{0.034}      \\
        Indoor\_26         & F-B Swinging             &                            & 1.694                                     & 6.120             & 7.098                     & 0.368            & 0.103             & \uline{0.034}  & \textbf{0.024}      \\
        Indoor\_27         & Turning                  &                            & 2.386                                     & 7.045             & $\times$                  & \uline{0.067}    & 0.169             & 0.451          & \textbf{0.029}      \\
        Indoor\_28         & F-B Swinging             &                            & 2.179                                     & 6.912             & $\times$                  & \uline{0.050}    & 0.088             & 8.192          & \textbf{0.022}      \\
        \bottomrule
        \end{tabular}
	\vspace{1pt}
	\begin{adjustwidth}{0cm}{0cm}
		\footnotesize{ Average depth refers to the mean distance of all points  in a frame. The units of Initial Velocity and Average Depth are $m/s$ and $m$, respectively. }
	\end{adjustwidth}
	\label{tab4}
\end{table*}

\section{EXPERIMENT RESULTS}
The performance of the proposed D-LI-Init method is experimentally evaluated on both open-source datasets and in real-world environments. All experiments are conducted on a laptop equipped with an Intel Core i7-9750H (2.60GHz) CPU.

\subsection{Public Dataset Validation}
We present the performance of the D-LI-Init method and LGO module on open-source datasets. Due to the current scarcity of LiDAR-inertial datasets with dynamic initialization, we evaluated the algorithm by starting SLAM at a specific point during the dataset sequences, simulating the initialization of SLAM while the system is in motion. 16 sequences from notable open-source datasets, including M2DGR \cite{M2DGR},  MCD \cite{MCD}, UrbanLoco \cite{9196526}, Newer Collegec \cite{ramezani2020newer}, Ground-challenge \cite{Ground-challenge}, and NTU-VIRAL \cite{nguyen2022ntu}, are selected and modified accordingly for this purpose. The chosen sequences encompass a variety of motion platforms, including handheld devices, vehicles, and UAV. They also cover diverse movement patterns such as vehicle rotation, high-speed turns, ascending and descending stairs with handheld devices, and gliding with UAV. These scenarios are selected to demonstrate the reliability and versatility of the proposed algorithm.

\begin{table*}
	\centering
	\caption{ATE (RMSE) in Meter Comparison of Real-World Proprietary Datasets (Outdoor)}
    \begin{tabular}{cccccccccc} 
    \toprule
    \textbf{Sequences} & \textbf{Motion Patterns} & \textbf{Platform}         & \textbf{Initial Velocity} & \textbf{Average Depth} & \textbf{FAST-LIO} & \textbf{LIO-SAM} & \textbf{KISS-ICP} & \textbf{I2EKF} & \textbf{D-LI-Init}  \\ 
    \midrule
    Outdoor\_1         & L-R Turns                & \multirow{21}{*}{Vehicle} & 1.214                     & 12.956                 & $\times$          & 1.918            & \uline{0.101}     & 2.933          & \textbf{0.012}      \\
    Outdoor\_2         & L-R Turns                &                           & 1.150                     & 13.363                 & $\times$          & 0.022            & 0.040             & \uline{0.020}  & \textbf{0.012}      \\
    Outdoor\_3         & Spinning                 &                           & 1.632                     & 14.087                 & $\times$          & 1.686            & \uline{0.049}     & 0.056          & \textbf{0.016}      \\
    Outdoor\_4         & Spinning                 &                           & 0.667                     & 6.927                  & 9.525             & \uline{0.041}    & 0.099             & 2.633          & \textbf{0.016}      \\
    Outdoor\_5         & Spinning                 &                           & 0.809                     & 20.941                 & 0.087             & \uline{0.028}    & 0.086             & 0.029          & \textbf{0.018}      \\
    Outdoor\_6         & L-R Turns                &                           & 1.079                     & 15.486                 & $\times$          & 4.143            & \uline{0.099}     & 0.165          & \textbf{0.041}      \\
    Outdoor\_7         & L-R Turns                &                           & 1.095                     & 19.901                 & $\times$          & 0.027            & 0.069             & \uline{0.038}  & \textbf{0.018}      \\
    Outdoor\_8         & Spinning                 &                           & 0.807                     & 20.840                 & 0.044             & \uline{0.019}    & 0.047             & 0.021          & \textbf{0.017}      \\
    Outdoor\_9         & L-R Turns                &                           & 1.500                     & 15.757                 & $\times$          & \uline{0.025}    & 0.068             & 0.395          & \textbf{0.015}      \\
    Outdoor\_10        & L-R Turns                &                           & 1.708                     & 14.256                 & $\times$          & 0.697            & \uline{0.074}     & 0.102          & \textbf{0.019}      \\
    Outdoor\_11        & Spinning                 &                           & 2.223                     & 13.339                 & $\times$          & 0.062            & 0.067             & \uline{0.041}  & \textbf{0.024}      \\
    Outdoor\_12        & L-R Turns                &                           & 1.405                     & 11.399                 & $\times$          & 0.659            & 0.062             & \uline{0.041}  & \textbf{0.025}      \\
    Outdoor\_13        & L-R Turns                &                           & 1.155                     & 14.871                 & $\times$          & \uline{0.034}    & 0.084             & 3.395          & \textbf{0.016}      \\
    Outdoor\_14        & Spinning                 &                           & 0.780                     & 19.474                 & $\times$          & \uline{0.027}    & 0.121             & 0.161          & \textbf{0.014}      \\
    Outdoor\_15        & L-R Turns                &                           & 1.280                     & 18.289                 & $\times$          & 5.134            & \uline{0.207}     & $\times$       & \textbf{0.020}      \\
    Outdoor\_16        & L-R Turns                &                           & 2.461                     & 6.636                  & $\times$          & 0.054            & 0.292             & \uline{0.049}  & \textbf{0.030}      \\
    Outdoor\_17        & L-R Turns                &                           & 1.041                     & 16.174                 & $\times$          & 0.686            & \uline{0.059}     & 0.922          & \textbf{0.018}      \\
    Outdoor\_18        & L-R Turns                &                           & 0.965                     & 14.128                 & $\times$          & 0.269            & \uline{0.091}     & 1.311          & \textbf{0.017}      \\
    Outdoor\_19        & L-R Turns                &                           & 2.669                     & 13.708                 & 1.856             & 0.044            & 0.058             & \uline{0.031}  & \textbf{0.020}      \\
    Outdoor\_20        & Spinning                 &                           & 2.173                     & 12.149                 & 3.447             & \uline{0.090}    & 0.127             & 0.097          & \textbf{0.090}      \\
    Outdoor\_21        & Spinning                 &                           & 0.891                     & 12.672                 & $\times$          & 0.455            & \uline{0.085}     & 0.299          & \textbf{0.021}      \\
    \bottomrule
    \end{tabular}
	\label{tab5}
    %\vspace{-12pt}
\end{table*}
\begin{figure}[t]
	\centering
	\includegraphics[width=0.9\linewidth]{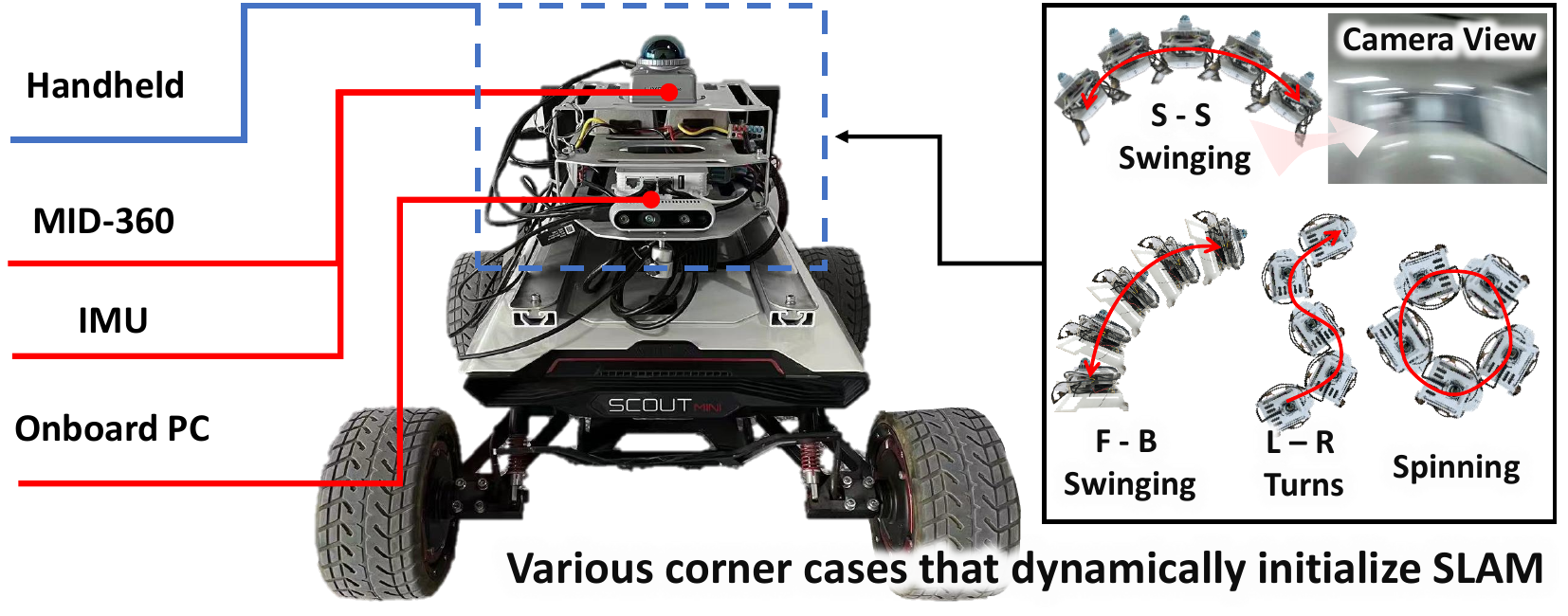}
	\caption{Vehicle-mounted and Handheld Devices.}
	\label{fig_6}
	%\vspace{-12pt}
\end{figure}

\subsubsection{Overall Performance}
The higher the accuracy of the initial state estimation, the faster and more precise the subsequent state estimation in LiDAR-inertial SLAM becomes. To quantitatively evaluate the quality of the initial state estimation, we integrated the D-LI-Init method into the FAST-LIO system for state estimation. We collected trajectory data for 10 seconds after initiating SLAM at a specific time period in the dataset (simulating the start of SLAM during motion) and analyzed the absolute trajectory error (ATE) \cite{evo} between the generated trajectories and the ground truth.

To highlight the accuracy and robustness of our method, we compared the results with those from FAST-LIO, LIO-SAM, KISS-ICP, CT-ICP, and I2EKF algorithms, as shown in Table \ref{tab1}. FAST-LIO represents tightly-coupled systems in a filter framework, while LIO-SAM serves as an example of a loosely-coupled system in an optimization framework. KISS-ICP, CT-ICP, and I2EKF are included as leading LiDAR-only odometry methods for comparison, evaluating the performance of the LiDAR-inertial SLAM systems enhanced by D-LI-Init.

\begin{table*}
	\centering
	\caption{The Performance of \textbf{Static} Initialization is Demonstrated through ATE (RMSE) in Meter Comparison.}
    \begin{tabular}{ccccccccc} 
    \toprule
    \textbf{Sequences} & \textbf{Motion Patterns} & \textbf{Platform}         & \textbf{ Average Depth} & \textbf{FAST-LIO} & \textbf{LIO-SAM} & \textbf{KISS-ICP} & \textbf{I2EKF} & \textbf{D-LI-Init}  \\ 
    \midrule
    Outdoor\_22        & \multirow{9}{*}{Static}  & \multirow{5}{*}{Vehicle}  & 13.248                  & \uline{0.00756}   & 0.03326          & 0.03918           & 0.02117        & \textbf{0.00645}    \\
    Outdoor\_23        &                          &                           & 15.557                  & \textbf{0.00789}  & 0.02622          & 0.03918           & 0.01940        & \uline{0.01284}     \\
    Outdoor\_24        &                          &                           & 10.934                  & \uline{0.00961}   & 0.03991          & 0.04774           & 0.01447        & \textbf{0.00809}    \\
    Outdoor\_25        &                          &                           & 19.027                  & \uline{0.01040}   & 0.03235          & 0.04378           & 0.02070        & \textbf{0.01034}    \\
    Indoor\_29         &                          &                           & 5.398                   & \uline{0.00612}   & 0.02422          & 0.03937           & 0.01602        & \textbf{0.00505}    \\ 
    \cmidrule{3-9}
    Indoor\_30         &                          & \multirow{4}{*}{Handheld} & 4.816                   & \textbf{0.00777}  & 0.04012          & 0.16229           & 0.02582        & \uline{0.01059}     \\
    Indoor\_31         &                          &                           & 6.681                   & \textbf{0.00690}  & 0.05171          & 0.09803           & 0.01958        & \uline{0.01681}     \\
    Indoor\_32         &                          &                           & 6.578                   & \textbf{0.00592}  & 0.02035          & 0.03821           & 0.02271        & \uline{0.00608}     \\
    Indoor\_33         &                          &                           & 7.278                   & \uline{0.00475}   & 0.05049          & 0.03927           & 0.02427        & \textbf{0.00309}    \\
%    \midrule
%    \multicolumn{4}{c}{Average}                                                                         & \textbf{0.00744}           & 0.03540          & 0.06078           & 0.02247        & \uline{0.00882}             \\
    \bottomrule
    \end{tabular}
	\label{tab6}
    %\vspace{-12pt}
\end{table*}

Experimental results demonstrate that tightly-coupled LiDAR-inertial SLAM systems like FAST-LIO have high demands for accurate initial states, and static initialization methods often fail to perform accurately or even lead to failure when SLAM is started during motion. In contrast, the LiDAR-inertial SLAM systems, enhanced with D-LI-Init, successfully perform dynamic initialization, ensuring accurate state estimation even when SLAM is initiated during motion. Moreover, the system shows high versatility, being applicable to various motion platforms and types. While the trajectory accuracy of the D-LI-Init enhanced system occasionally slightly lags behind that of CT-ICP, it generally outperforms other methods, ensuring reliable initialization in dynamic scenarios.

\subsubsection{LGO Performance}
The performance of the LGO module further affects the accuracy of the initial value estimation. To evaluate LGO module, nine sequences are selected without any special preprocessing, and the ATE between the trajectory of algorithm's output and the ground truth is calculated. To emphasize the accuracy of the LGO module, the results are compared with those of the FAST-LIO, KISS-ICP, and CT-ICP algorithms, as shown in Table \ref{tab2}. The corresponding trajectory are shown in Fig. \ref{fig_5}. The experimental results demonstrate that the overall performance of LGO surpasses that of LO and is comparable to, or in some cases, even better than LIO.

\subsubsection{Initial Values Recovery}
To evaluate the accuracy of the initial velocity estimation, the ground truth velocity is not available in open-source datasets, making direct comparison impossible. 
Considering that FAST-LIO provides accurate state estimates when initialized from a stationary state, we chose the starting point of the open-source dataset sequence (where the device was stationary) to perform state estimation using FAST-LIO. The estimated velocity at a specific time point, simulating the dynamic initialization of SLAM, is then used as the ground truth for initial velocity. Given that the true velocity and the estimated initial velocity may differ in direction due to being referenced in different coordinate systems, the magnitude of both velocities is compared as the standard for evaluating the initial velocity. The specific results are presented in Table \ref{tab3}. The results indicate that the maximum initial velocity error is 1 $m/s^2$, with the overall error being relatively small. This demonstrates that the calculated initial velocity is highly accurate and can serve as a high-quality initial value for subsequent LIO.

To assess the effectiveness of our method in estimating the gravity vector and initial gyroscope biases, we collected further refined gyroscope bias and gravity vector data from LiDAR-inertial SLAM systems initialized using D-LI-Init. The accuracy of our method is indirectly assessed by comparing the differences between these refined values and the initial values provided by D-LI-Init. As shown in Fig. \ref{fig_4}, the subsequent refinement of the initial gyroscope biases and gravity vector is minimal, indicating that the initial values calculated by our method are highly accurate.

\subsection{Real-world Testing}
Ensuring the reliability of the algorithm in real-world scenarios is crucial. Data collection is conducted using handheld and vehicle-mounted devices in both indoor and outdoor environments, with data gathering initiated during device motion and continuing for 10 seconds, as shown in Fig. \ref{fig_6}. It is important to emphasize that during this process, high-quality undistorted point clouds are matched against prior maps to obtain the global pose of the current frame, which served as the ground truth trajectory. This trajectory is used to accurately evaluate the performance of the D-LI-Init method. Additionally, the corresponding experimental dataset will be made open-source, marking the first open-source dataset specifically designed to evaluate the initialization capabilities of LIO.

\subsubsection{Indoor Environments}
The indoor environment used for testing covers an area of over 1500 square meters on the first floor of an experimental building. The specific results are shown in Table \ref{tab4}. It is demonstrated that the original static initialization module of FAST-LIO is inadequate for dynamic initialization, often leading to system failure and preventing successful startup. In rare cases where initialization is completed, subsequent state estimation exhibited significant errors due to inaccurate initial estimates. In contrast, the D-LI-Init method is able to reliably estimate initial values under various dynamic conditions, ensuring that the system maintained high accuracy in state estimation. Overall, as the initial velocity increases, dynamic initialization becomes more challenging, leading to a slight decrease in accuracy.

\subsubsection{Outdoor Environments}
The effectiveness of the proposed D-LI-Init method is also evaluated in an outdoor environment. The data collection vehicle, as shown in Fig. \ref{fig_6}, is used to gather data around a building on a university campus. The specific results are presented in Table \ref{tab5}. These results demonstrate that the inertial system, enhanced by the D-LI-Init method, can initiate SLAM under various outdoor motion conditions with accurate initial values, ensuring that the system maintains an accuracy within 0.09 meters throughout subsequent operations. By analyzing the data from Tables \ref{tab4} and \ref{tab5} it can be observed that as the average depth of the LiDAR-perceived scene increases, the system's robustness improves, and accuracy slightly increases.

\begin{table}
	\centering
	\caption{The Time Consumed by Each Module (ms)}
    \begin{tabular}{cccccc} 
    \toprule
    \textbf{Sequences} & \textbf{LGO} & \textbf{LIO} & \textbf{Alignment} & \textbf{Iter.} & \textbf{Ave.}  \\ 
    \midrule
    Indoor\_1          & 33.42        & 41.94        & 0.0079          & 2              & 109.23        \\
    Indoor\_4          & 27.18        & 36.19        & 0.0068          & 2              & 99.52        \\
    Indoor\_15         & 30.15        & 41.12        & 0.0072          & 1              & 70.45        \\
    Outdoor\_1         & 32.53        & 40.04        & 0.0073          & 3              & 143.26        \\
    \bottomrule
    \end{tabular}
	\vspace{1pt}
	\begin{adjustwidth}{0cm}{0cm}
		\footnotesize{ ``Iter.'' denotes Iteration number, ``Ave.'' denotes average time per frame. The LGO and LIO modules measure the average time per frame, while the alignment module measures the average time per alignment.} 
	\end{adjustwidth}
	\label{tab7}
    %\vspace{-12pt}
\end{table}

\subsubsection{Stationary State Initialization Performance}
Simultaneously, a comparison is conducted between the D-LI-Init method and the stationary initialization method under static conditions, as presented in Table \ref{tab6}. The results of both methods are found to be very close, indicating that the D-LI-Init method is also capable of accurately solving for initial values under stationary conditions, demonstrating its robustness irrespective of the specific motion pattern.

\subsection{Time Consumption Evaluation}
To evaluate the time efficiency of the proposed D-LI-Init method, a series of timing tests are conducted on randomly selected sequences, as shown in Table \ref{tab7}. The results indicate that the majority of the processing time is concentrated in the LiDAR odometry module, while the time consumed by the alignment module is negligible. Therefore, the real-time performance of the LGO module is of paramount importance. The single-frame processing time for the LiDAR odometry module (LGO and LIO) can be maintained at approximately 50ms, which is within an acceptable range. However, the motion state at the initial moment influences the number of iterations required by the D-LI-Init method, potentially increasing the processing time. Considering the total time required for odometry calculation using the 20 frames of data collected during the initialization process is typically around 1.5 seconds, a maximum iteration limit of three is set to ensure that the method's processing time remains within 3 seconds. This constraint is intended to keep the maximum processing time close to the actual data collection time, thereby ensuring the real-time performance of the initialization module.

\section{Conclusion}
In this paper, we propose a dynamic initialization method for LiDAR-inertial systems. It has been demonstrated to be applicable across various platforms, providing accurate initial values without being constrained by specific motion patterns. 
%Additionally, we propose the LGO module, which offers an alternative sensor fusion strategy for LiDAR-inertial systems when encountering system faults. 
Notably, this method is highly dependent on the accuracy of the LiDAR odometry, and we will conduct further research in the area of enhancing LiDAR odometry accuracy.

\bibliographystyle{IEEEtran}
\bibliography{ref}

\vfill

\end{document}